\documentclass{article}

\usepackage{microtype}
\usepackage{graphicx}
\usepackage{subfigure}
\usepackage{booktabs} 
\usepackage{lipsum}
\usepackage{multirow}

\usepackage{hyperref}

\usepackage[accepted]{icml2025}

\usepackage{amsmath}
\usepackage{amssymb}
\usepackage{mathtools}
\usepackage{amsthm}

\usepackage[capitalize,noabbrev]{cleveref}

\theoremstyle{plain}

\theoremstyle{definition}

\theoremstyle{remark}

\usepackage[textsize=tiny]{todonotes}

\icmltitlerunning{Leveraging the Structure of Medical Data for Improved Representation Learning}

\begin{document}

\twocolumn[
\icmltitle{Leveraging the Structure of Medical Data for Improved Representation Learning}



\icmlsetsymbol{equal}{*}
\icmlsetsymbol{equal_last}{†}
\begin{icmlauthorlist}
\icmlauthor{Andrea Agostini}{equal,yyy}
\icmlauthor{Sonia Laguna}{equal,yyy}
\icmlauthor{Alain Ryser}{equal,yyy}
\icmlauthor{Samuel Ruiperez-Campillo}{equal,yyy}
\icmlauthor{Moritz Vandenhirtz}{yyy}
\icmlauthor{Nicolas Deperrois}{comp}
\icmlauthor{Farhad Nooralahzadeh}{comp}
\icmlauthor{Michael Krauthammer}{comp}
\icmlauthor{Thomas M. Sutter}{equal_last,yyy}
\icmlauthor{Julia E. Vogt}{equal_last,yyy}
\end{icmlauthorlist}

\icmlaffiliation{yyy}{Department of Computer Science, ETH Zurich, Zurich, Switzerland}
\icmlaffiliation{comp}{Department of Quantitative Biomedicine, University of Zurich, Zurich, Switzerland}

\icmlcorrespondingauthor{}{\{anandrea, slaguna, aryser, sruiperez\}@inf.ethz.ch}

\icmlkeywords{Machine Learning, ICML}

\vskip 0.3in
]



\printAffiliationsAndNotice{\icmlEqualContribution \icmlEqualLastContribution} 

\begin{abstract}
Building generalizable medical AI systems requires pretraining strategies that are data-efficient and domain-aware. Unlike internet-scale corpora, clinical datasets such as MIMIC-CXR offer limited image counts and scarce annotations, but exhibit rich internal structure through multi-view imaging.
We propose a self-supervised framework that leverages the inherent structure of medical datasets.
Specifically, we treat paired chest X-rays (i.e., frontal and lateral views) as natural positive pairs, learning to reconstruct each view from sparse patches while aligning their latent embeddings. Our method requires no textual supervision and produces informative representations. Evaluated on MIMIC-CXR, we show strong performance compared to supervised objectives and baselines being trained without leveraging structure.
This work provides a lightweight, modality-agnostic blueprint for domain-specific pretraining where data is structured but scarce.
\end{abstract}

\section{Introduction}

Advances in foundation models have ignited interest in clinical AI, yet the data realities of medicine differ sharply from the internet-available corpora that power vision-language behemoths. Within the medical domain, in the popular field of radiology, chest-X-ray collections such as MIMIC-CXR comprise hundreds of thousands—not billions—of images, and every expert annotation carries substantial monetary and time cost \citep{hassanzadeh2018clinical}. Consequently, frontier general-purpose models still lag behind experts on domain benchmarks and remain costly to fine-tune or deploy in practice \citep{chaves2024towards}. To learn informative representations without incurring those expensive medical labels, self-supervised learning has emerged as a useful approach \citep{tiu2022expert, azizi2021big}. Among the many approaches, domain-adapted masked autoencoders, which reconstruct held-out patches, and complementary contrastive objectives that distinguish latent views have proven effective for distilling rich features from unlabeled exams \citep{xiao2023delving}. Importantly, these techniques are most effective when the pre-text task, visible-patch ratio, and augmentations are re-engineered around medical imagery's limited scale and anatomical regularities \citep{mo2024multimed}.

Beyond clever objectives, the latent organization of clinical datasets is an under-explored source of supervision. Medical studies frequently present multi-view pairs, longitudinal follow-ups, and paired image–report examples, all of which encode consistent anatomy or semantics across views. Exploiting such coherence, whether through multi-view contrast, cross-modal alignment, or multitask training, has been shown to improve robustness and label efficiency in recent multimodal radiology models \citep{mo2024multimed,pellegrini2025radialog,chen2401chexagent}. Therefore, capitalizing on these built-in redundancies offers a scalable path toward domain-specific foundation models. 

The present work follows this intuition, proposing a multi-view regularized masked autoencoder framework that leverages paired chest-X-ray views to learn view-invariant representations without relying on external supervision. Unlike recent vision–language models that depend on paired reports \citep{pellegrini2025radialog, chen2401chexagent}, our method exploits the complementary information provided by multiple X-ray views acquired during a patient examination. Concretely, we treat the frontal–lateral views in each MIMIC-CXR exam as natural positives. For every view, we run a masked autoencoder that reconstructs its own missing patches, regularizing the latent space. In a separate head we apply a regularization objective, e.g. a contrastive loss that pulls together the frontal and lateral embeddings of the same study while pushing apart embeddings from different samples. Optimized jointly, these complementary objectives yield features that are simultaneously detail-rich and view-invariant, with no textual supervision or annotation burdens of image–report pairs. Although we focus on chest radiographs, the same strategy could extend to any life-science domain that offers repeated or complementary scans (e.g., longitudinal MRIs or multi-sequence CT).

Our study makes two principal contributions: (i) We show that explicitly exploiting the internal structure of clinical datasets enables effective pretraining on limited data, outperforming supervised training on downstream tasks; (ii) we design the first multi-view MAE and contrastive pipeline for radiology and demonstrate consistent gains on MIMIC-CXR generalizing across modalities, establishing a lightweight, modality-agnostic blueprint for future medical foundation models.

\section{Dataset and Preprocessing}
\label{sec:dataset}
\begin{figure}
    \centering
    \includegraphics[width=0.48\textwidth]{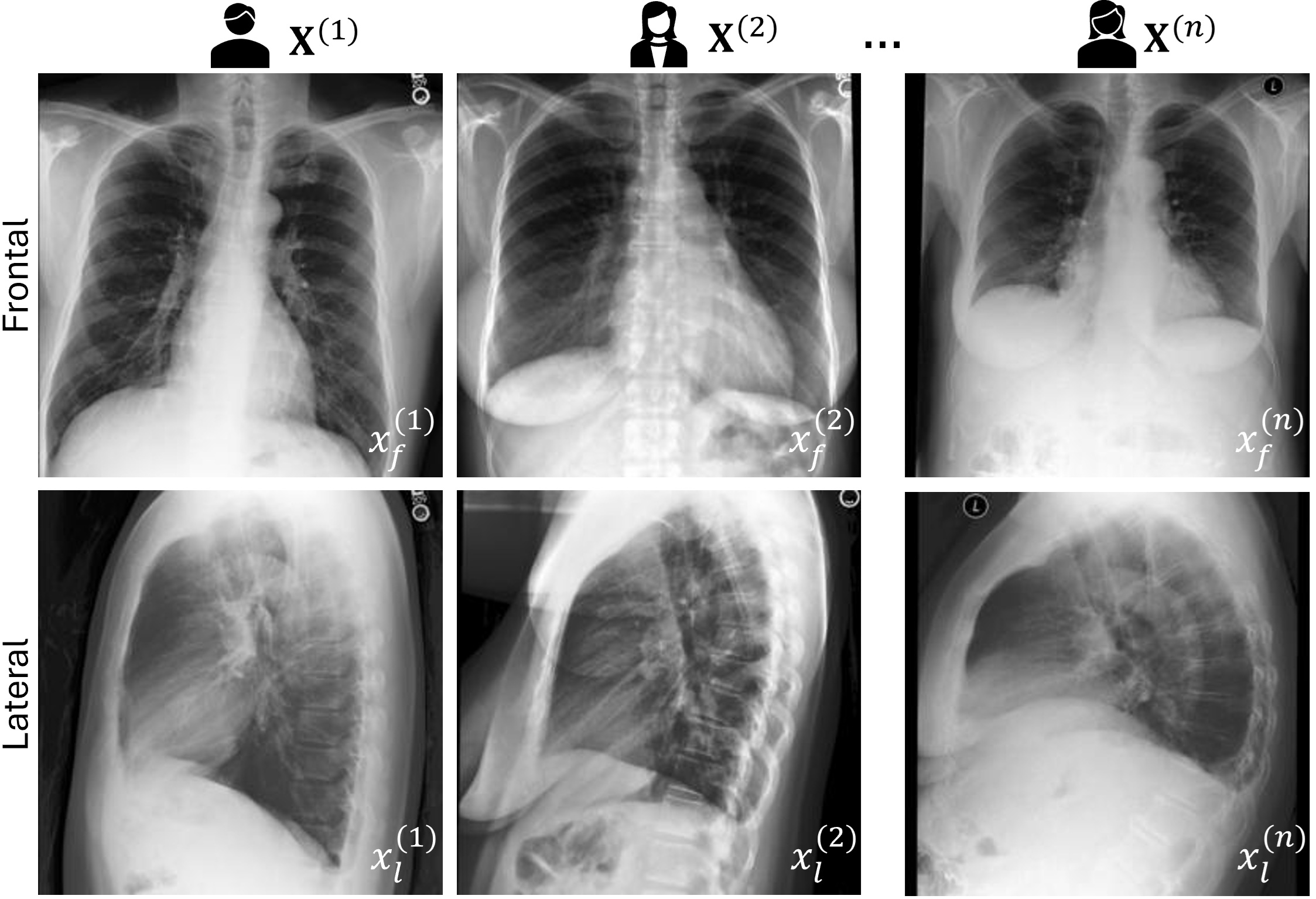}
    \caption{Frontal–lateral pairs from MIMIC-CXR.  
    Each column shows a frontal image ($\mathbf{x}^{(i)}_f$, top) and its matching lateral view ($\mathbf{x}^{(i)}_l$, bottom).
    }
    \vspace{-0.5cm}
    \label{fig:dataset}
\end{figure}

\paragraph{MIMIC-CXR as a bimodal resource.}
We conduct all experiments on the publicly available \emph{MIMIC-CXR} archive \citep{johnson2019mimic}, a large-scale collection of routine chest radiographs acquired in critical-care settings. Each imaging \emph{study} bundles every projection that shares one radiology report and fourteen diagnostic labels derived by CheXpert \citep{irvin2019chexpert}. Image quality varies markedly because of patient positioning, bedside hardware, and emergent clinical constraints \citep{raoof2012interpretation}.  
To expose inherent view redundancy, we isolate two complementary projection families—\textit{frontal} (posterior–anterior \textit{PA} or anterior–posterior \textit{AP}) and \textit{lateral} (left-lateral \textit{LL} or generic \textit{Lateral}). Whenever a study contains at least one image from each family, we enumerate every frontal–lateral combination to create paired samples (see \cref{fig:dataset}), yielding a collection $\mathcal{X} = \{\mathbf{X}^{(i)}\}_{i=1}^{N}$ with $\mathbf{X}^{(i)} = \{\mathbf{x}^{(i)}_{f},\,\mathbf{x}^{(i)}_{l}\}$, where $\mathbf{x}^{(i)}_{f}$ and $\mathbf{x}^{(i)}_{l}$ are a frontal and a lateral radiograph from the $i^{th}$ study and $\mathcal{V} = \{ f, l \}$ defines the set of views. A raw image may appear in multiple tuples, yet every pair is unique by construction. This framing supplies natural positive pairs for self-supervision while preserving the study-centric semantics of the original dataset.

\paragraph{Preprocessing pipeline and data split.}
Each radiograph is center-cropped and isotropically resized to $224\!\times\!224$. Intensities are rescaled to $[0,1]$ and standardized with ImageNet statistics. Study-level labels are inherited from the MIMIC-CXR-JPG release \citep{johnson2024mimic}, produced by the CheXpert labeler. Following \citet{Haque2021.07.30.21261225}, the three non-positive states (“negative”, “not mentioned”, “uncertain”) are collapsed into a single $0$-class, treating only explicit positives as $1$. To prevent patient leakage, we partition \emph{subjects}—and thus all associated studies and image pairs—into training ($80\%$), validation ($10\%$), and test ($10\%$) splits.

\section{Method}

\begin{figure*}
    \centering
    \includegraphics[width=\textwidth]{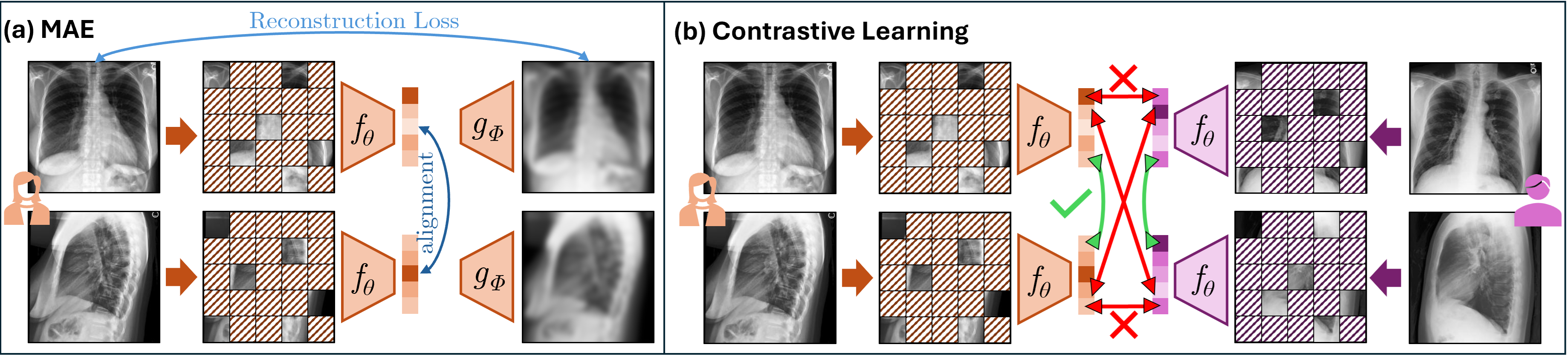}
    \caption{Pretraining strategies for multi-view chest radiographs. \textbf{(a)} \textit{MAEs} reconstruct masked patches from visible ones using an encoder–decoder architecture and optionally enforce alignment between frontal and lateral views. \textbf{(b)} \textit{Contrastive Learning} maximizes agreement between corresponding views in the same study while contrasting against other samples in the batch using a contrastive loss.}
    \label{fig:pretraining}
\end{figure*}

We compare two different pretraining paradigms in this work.
The first method combines a reconstruction loss with an additional alignment loss between views, where the second approach applies a multi-view contrastive learning approach \citep{tian2020contrastive}.
An overview of the two pretraining paradigms is shown in \cref{fig:pretraining}.

\paragraph{Multiview MAE} 
MAEs \citep{he2022masked} are a self-supervised learning approach designed to learn high-quality representations by reconstructing missing portions of the input. MAEs randomly mask a large fraction of the input data and train an encoder-decoder architecture to recover the masked content from the visible subset.
In this work, we assume the encoder and decoder to be vision transformers \citep[ViTs,][]{dosovitskiy2020image}.

The encoder $f_\theta$ processes only the unmasked input tokens $\mathbf{x}_{\text{vis}}^{(i)} = (1 - M(\mathbf{x}^{(i)}))$ producing latent representations $\mathbf{z}^{(i)} = f_\theta(x_{\text{vis}}^{(i)})$. $M(\cdot)$ applies a random mask to the input to mask. These are passed, along with mask token placeholders, to a lightweight decoder $g_\phi$, which attempts to reconstruct the original input $x$, including the masked parts. The model is trained by minimizing a reconstruction loss $\mathcal{L}_{\text{Rec}}$ over only the masked positions:
\begin{align}
\mathcal{L}_{\text{Rec}}(\mathbf{x}^{(i)}) \propto \sum_{t \in M(\mathbf{x}^{(i)})} \left\| \mathbf{x}^{(i)}_t - g_\phi(f_\theta(\mathbf{x}^{(i)}_{\text{vis}}))_t \right\|_2^2 \nonumber.
\end{align}
Here, $\mathbf{x}^{(i)}_t$ denotes the original input at masked position $t$, and $g_\phi(f_\theta(\mathbf{x}^{(i)}_{\text{vis}}))_t$ is the decoder's reconstruction at that position with $\mathbf{x}^{(i)}_{\text{vis}}$ being the visible or un-masked part of the input image.
For more details, we refer to \citet{he2022masked}.

We propose a late fusion approach to adapt the MAE objective to the multi-view setting.
In addition to the reconstruction, we emphasize similarity between the latent representations through an additional alignment objective $\mathcal{L}_{\text{Align}}$ \citep{sutter2024unity,agostini2024weakly}.

\begin{align}
    \mathcal{L}_{\text{Align}}(\mathbf{x}^{(i)}_f, \mathbf{x}^{(i)}_l) \propto \sum_{t=1}^T d_{\text{MSE}}(f_\theta(\mathbf{x}^{(i)}_f)_t, f_\theta(\mathbf{x}^{(i)}_l)_t),
\end{align}
where $d_{\text{MSE}}(\cdot, \cdot)$ is the mean squared error (MSE).
In this work, we only consider the MSE as an alignment metric, but the extension to other measures is part of future work.

The objective of the proposed multi-view MAE approach, MVMAE, for a pair of images $x_f$ and $x_l$ follows as
\begin{align}
    \mathcal{L}(\mathbf{x}^{(i)}_f, \mathbf{x}^{(i)}_l) = \frac{1}{|\mathcal{V}|} \sum_{v \in \mathcal{V}} \mathcal{L}_{\text{Rec}}(\mathbf{x}^{(i)}_v) + \beta \cdot \mathcal{L}_{\text{Align}}(\mathbf{x}^{(i)}_f, \mathbf{x}^{(i)}_l), \nonumber
\end{align}
where $\beta$ is an additional weighting between reconstruction and alignment loss.

\paragraph{Contrastive Learning} 
To encourage view-invariant representations across paired frontal and lateral chest radiographs, we adopt a contrastive learning objective inspired by SimCLR~\citep{chen2020simple}. For each image pair $\mathbf{X}^{(i)} = \{\mathbf{x}^{(i)}_f, \mathbf{x}^{(i)}_l \}$ in a study, we encode both views using a shared transformer encoder backbone and extract latent representations from the [CLS] token, as used in~\citet{radford2021learning}. The contrastive loss is computed as:


\begin{equation}
\mathcal{L}_{\mathrm{Con}}(\mathcal{X}) = -\frac{1}{2N} \sum_{\substack{v \in \mathcal{V}}} \sum_{i=1}^{N} \Gamma(\mathcal{X}, v, i),
\end{equation}

\noindent where
\[
\resizebox{\linewidth}{!}{$
\Gamma(\mathcal{X}, v, i) = \log \left( 
\frac{\Lambda(\mathbf{x}^{(i)}_v, \mathbf{x}^{(i)}_{\bar{v}})}
{
\sum_{\substack{k=1 \\ k \ne i}}^{N} \Lambda(\mathbf{x}^{(i)}_v, \mathbf{x}^{(k)}_v) 
+ 
\Lambda(\mathbf{x}^{(i)}_v, \mathbf{x}^{(k)}_{\bar{v}})
} \right),
$}
\]

\noindent where \(\Lambda(\mathbf{x}^{(i)}_v, \mathbf{x}^{(i)}_{\bar{v}}) = \exp\left( \mathrm{sim}(f_\theta(\mathbf{x}^{(i)}_v), f_\theta(\mathbf{x}^{(i)}_{\bar{v}}))/\tau \right)\), and \(\mathrm{sim}(\mathbf{u}, \mathbf{w}) = \mathbf{u}^\top \mathbf{w}\) denotes the cosine similarity between the $\ell_2$-normalized vectors \(\mathbf{u}\) and \(\mathbf{w}\), \(\tau\) is a temperature parameter, and $\bar{v}$ defines a view $\neq v$.
The pair of normalized embeddings $(\mathbf{z}_{f}^{(i)}, \mathbf{z}_{l}^{(i)})$ from the same study $i$ is treated as a positive pair where $\mathbf{z}_{v}^{(i)} = f_\theta(\mathbf{x}^{(i)}_v)$. Embeddings from different studies $j \neq i$ regardless of the view (i.e., $\mathbf{z}_{f}^{(j)}$ and $\mathbf{z}_{l}^{(j)}$) serve as negative samples.
In distributed training, negatives are gathered across devices for a more expressive representation without extra annotation burden. 

\vspace{-0.13cm}
\section{Experiments}
    
\begin{figure*}
    \centering
    \includegraphics[width=\textwidth]{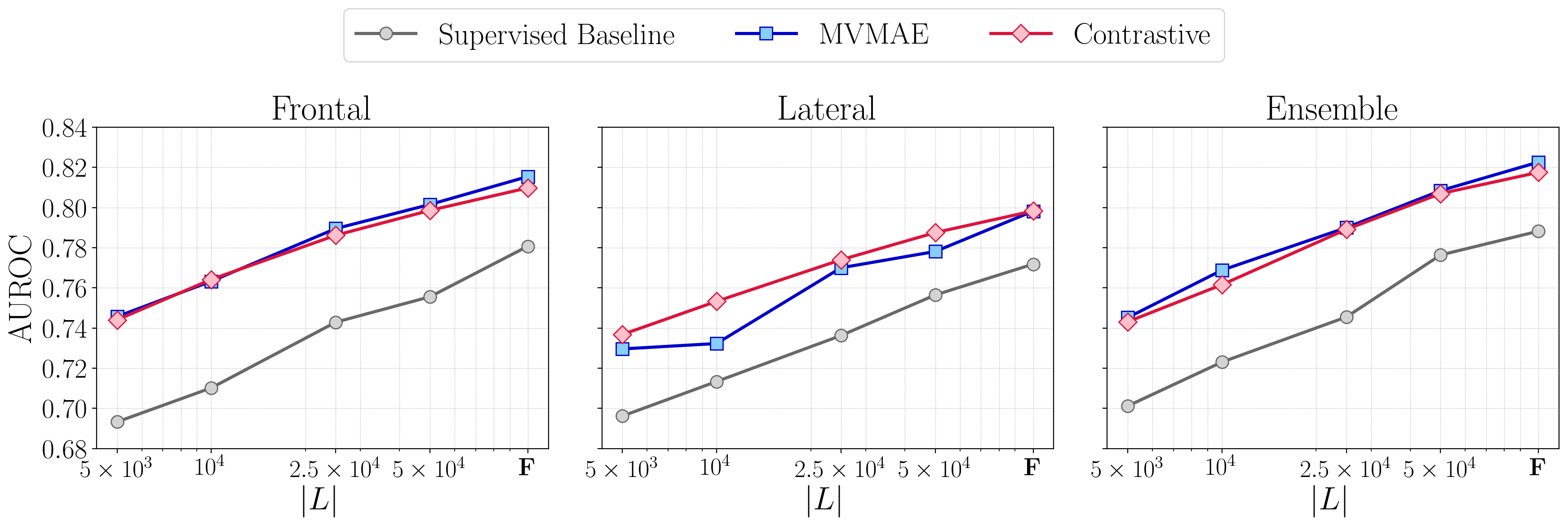}
   \caption{
Performance comparison of MVMAE, Contrastive, and Supervised methods across three evaluation settings: (a) \textit{Frontal}, (b) \textit{Lateral}, and (c) \textit{Ensemble}. Each plot shows the AUROC score, computed as a macro-average across 14 pathology labels, achieved by each method under varying numbers of labeled samples $|L|$. \textbf{F} denotes the total dataset size (10.2$\times10^4$).
}
\vspace{-0.2cm}
    \label{fig:comparison}
\end{figure*}


We aimed to investigate how the representations learned through our MVMAE models and the contrastive learning approach influence performance on a downstream classification task under varying levels of label availability.  This setting mimics realistic clinical practice, where expert annotation is scarce and expensive.  We study two complementary questions: (i) \emph{How effective are the representations learned by a pretrained encoder when the model is fully fine-tuned on a downstream classification task?} In particular, we assess whether leveraging the data structure during pretraining leads to measurable improvements. (ii) \emph{To what extent is fine-tuning necessary? Can we reach comparable performance via linear probing?} For this, we would keep the encoder frozen and train only a linear classifier.



We compare three training regimes that share the same ViT-b backbone \citep{dosovitskiy2020image} and optimizer but differ in how that backbone is initialized: the proposed MVMAE; the proposed contrastive-only variant that drops the reconstruction loss; a purely supervised baseline trained from scratch.
All self-supervised models are first exposed to the full unlabeled training pool; downstream experiments are then performed on splits of 5K, 10K, 25K, 50K, and the complete 102K labeled studies drawn from the training set. We then compute the numbers over the full validation set.
All models were trained using data augmentations common in this type of learning paradigm: random resize cropping, horizontal clipping, and jitter addition to improve the prediction and representation capabilities.
We perform this analysis through two experiments. 

\textbf{Experiment 1} assesses \emph{label-efficiency under end-to-end fine-tuning}.  After attaching a randomly initialized linear classification head, we unfreeze all the encoder weights and train them on each supervision tier, keeping learning-rate schedules identical so that any performance difference can be attributed to the representation rather than to additional compute.  Monitoring the AUROC (macro-averaged over all labels) as the number of labels grows allows us to quantify how quickly each pretraining strategy narrows the gap to its fully supervised counterpart. Results are reported in \cref{fig:comparison} evaluated across three modality scenarios: the set of \textit{Frontal} views, \textit{Lateral} views, and an \textit{Ensemble} of both.

\textbf{Experiment 2} probes \emph{representation quality with the encoder frozen}.  Here, we restrict supervision to the smallest 5K-label subset and compare the full fine-tuning approach from Experiment 1 to a linear probing approach with only a trainable linear classifier head. Because encoder weights remain fixed in the latter, any improvement must originate from the quality of the features rather than from further optimization of the backbone. Results are summarized in \cref{table:strategy-type-comparison} right panel.

\begin{table}[t]
\centering
\footnotesize
\renewcommand{\arraystretch}{1.0} 
\begin{tabular}{l l c c}
\toprule
\textbf{Strategy} & \textbf{Modality} & \textbf{Fine-tuning} & \textbf{Linear Probing} \\ \midrule
\multirow{3}{*}{Supervised} 
  & Frontal  & 0.69 & - \\
  & Lateral  & 0.70 & - \\
  & Ensemble & 0.70 & - \\ \midrule
\multirow{3}{*}{MVMAE} 
  & Frontal  &  0.75 & 0.65 \\
  & Lateral  & 0.73 & 0.65 \\
  & Ensemble & 0.75 & 0.69 \\ \midrule
\multirow{3}{*}{Contrastive} 
  & Frontal  & 0.74 & 0.65 \\
  & Lateral  & 0.74 & 0.64 \\
  & Ensemble & 0.74 & 0.66 \\ 
\bottomrule
\end{tabular}
\caption{\textbf{Comparison of fine-tuning the pretrained encoder vs linear probing across methods and modality types on the classification task.} Each method (Supervised, MVMAE, Contrastive) is evaluated using different types (Frontal, Lateral, Ensemble) and trained only on 5000 labeled samples.}
\vspace{-0.6cm}
\label{table:strategy-type-comparison}

\end{table}

Note that the ensemble is built as a late fusion of the unimodal single view scores. All models are selected on a held-out validation set and reported on the test set using macro-AUROC over the fourteen CheXpert labels. 

\vspace{-0.3cm}
\paragraph{Discussion} From our results, we see that
(1) Multimodal pretraining consistently improves downstream classification performance compared to training from scratch.
(2) Leveraging data structure during pretraining proves more effective than enforcing structure at the supervision stage only (i.e., via an ensemble supervised classifier). Notably, unimodal performance of the pretrained models exceeds or matches that of the supervised ensemble classifier, suggesting that soft information sharing during pretraining is an effective way to exploit the underlying data structure. (3) Fine-tuning the encoders previously learnt proves more useful than directly probing the representations.



\section{Conclusion}
We present a multi-view regularized masked autoencoder that leverages the natural structure of clinical imaging data—specifically, paired anatomical views—to learn robust, informative representations without requiring textual supervision. By combining masked reconstruction with cross-view alignment, our approach enables efficient pretraining leveraging the inherent structure of medical datasets.
Experiments on MIMIC-CXR demonstrate that this strategy closes much of the gap to vision–language models while remaining lightweight and domain-adaptable. Looking forward, the framework is broadly applicable to other structured medical modalities, such as longitudinal studies or multi-sequence scans, offering a scalable path toward foundation models grounded in the realities of clinical data.

\section*{Acknowledgments}

This work was supported as part of the Swiss AI Initiative by a grant from the Swiss National Supercomputing
Centre (CSCS) under project ID a02 on Alps, and by the LOOP Zurich as part of the application driver project supporting the LOOP Zurich Biomedical Informatics Platform (BMIP). TS and AA are supported by the grant \#2021-911 of the Strategic Focal Area “Personalized Health and Related Technologies (PHRT)” of the ETH Domain (Swiss Federal Institutes of Technology). MV and SL are supported by the Swiss State Secretariat for Education, Research, and Innovation (SERI) under contract number MB22.00047. AR is supported by the StimuLoop grant \#1-007811-002 and the Vontobel Foundation. ND and FN received research support from the Digitalization Initiative of the Zurich Higher Education Institutions (DIZH)- Rapid Action Call - under TRUST-RAD project. MK is supported by the UZH Global Strategy and Partnerships Funding Scheme and a Research Partnership Grant with China, Japan, South Korea and the ASEAN region (RPG 072023 18).

\bibliography{example_paper}
\bibliographystyle{icml2025}





\end{document}